\documentclass[letterpaper, 10 pt, journal, twoside]{IEEEtran}  

\usepackage{graphics} 
\usepackage{epsfig} 
\usepackage{amsmath} 
\usepackage{amssymb}  
\usepackage{amsfonts}
\usepackage{multicol}
\usepackage{multirow}
\usepackage[bookmarks=true]{hyperref}
\usepackage{optidef}
\usepackage{comment}
\usepackage[dvipsnames]{xcolor}
\usepackage[noline,ruled,linesnumbered]{algorithm2e}
\usepackage[capitalize]{cleveref}
\usepackage{xurl}
\usepackage{everysel}
\usepackage{cite}

\newcommand{\exppol}{\pi_{\mathrm{exp}}}

\newcommand{\opt}{\mathrm{Opt}}
\newcommand{\initcond}{\sigma_0}
\newcommand{\traj}{\tau}
\newcommand{\olloss}{\mathcal{L}}
\newcommand{\dagname}{\text{DA\footnotesize{GGER}}}

\newcommand{\ol}{\mathrm{ol}}
\newcommand{\ft}{\mathrm{ft}}
\newcommand{\start}{\mathrm{start}}
\newcommand{\goal}{\mathrm{goal}}

\newcommand{\dynapprox}{\hat{f}}

\newcommand{\cost}{J}
\newcommand{\terminalcost}{J_{\mathrm{T}}}
\newcommand{\statedim}{n_x}
\newcommand{\controldim}{n_u}

\newcommand{\bx}{\mathbf{x}}
\newcommand{\bX}{\mathbf{X}}
\newcommand{\bu}{\mathbf{u}}
\newcommand{\bU}{\mathbf{U}}

\newcommand{\bV}{\mathbf{V}}
\newcommand{\boldf}{\mathbf{f}}

\newcommand{\boe}{\textbf{\oe}}
\newcommand{\br}{\mathbf{r}}
\newcommand{\bv}{\mathbf{v}}
\newcommand{\bT}{\mathbf{T}}
\newcommand{\horizon}{N}
\newcommand{\rhchorizon}{h+H}
\newcommand{\params}{\theta}

\newcommand{\calJ}{\mathcal{J}}
\newcommand{\calT}{\mathcal{T}}
\newcommand{\calR}{\mathcal{R}}
\newcommand{\calC}{\mathcal{C}}
\newcommand{\calX}{\mathcal{X}}
\newcommand{\calU}{\mathcal{U}}
\newcommand{\calD}{\mathcal{D}}
\newcommand{\calH}{\mathcal{H}}

\newcommand{\real}{\mathbb{R}}
\newcommand{\integer}{\mathbb{N}}

\newcommand{\red}{\textcolor{red}}

\newcommand{\black}{\textcolor{black}}

\renewcommand{\baselinestretch}{0.94} 
\usepackage[font=small,skip=4pt]{caption}
\EverySelectfont{%
\fontdimen2\font=0.2em
\fontdimen3\font=0.1em
\fontdimen4\font=0.07em
}

\title{ Transformer-based Model Predictive Control: \\ Trajectory Optimization via Sequence Modeling
}

\author{Davide Celestini$^{1*}$, Daniele Gammelli$^{2*}$, Tommaso Guffanti$^{2}$, Simone D'Amico$^{2}$,\\ Elisa Capello$^{1}$, and Marco Pavone$^{2}$
\thanks{Manuscript received: May, 16, 2024; Revised August, 4, 2024; Accepted September, 3, 2024. This letter was recommended for publication by Editor J. Kober upon evaluation of the reviewers’ comments.
This work is supported by Blue Origin (SPO $\#$299266) as Associate Member and Co-Founder of the Stanford’s Center of AEroSpace Autonomy Research (CAESAR) and the NASA University Leadership Initiative (grant $\#$80NSSC20M0163). This article solely reflects the opinions and conclusions of its authors and not any Blue Origin or NASA entity.
\textit{(Corresponding author: Davide Celestini.)}} 
\thanks{$^{*}$ Equal contribution.}
\thanks{$^{1}$ Davide Celestini and Elisa Capello are with the Department of Mechanical and Aerospace Engineering, Politecnico di Torino, 10129 Torino, Italy. (e-mail : davide.celestini@polito.it,  elisa.capello@polito.it)}%
\thanks{$^{2}$ Daniele Gammelli, Tommaso Guffanti, Simone D'Amico and Marco Pavone are with the Department of Aeronautics and Astronautics, Stanford University, 94305, USA. (e-mail : gammelli@stanford.edu,  tommaso@stanford.edu, damicos@stanford.edu, pavone@stanford.edu)}%
\thanks{Digital Object Identifier (DOI): \href{https://doi.org/10.1109/LRA.2024.3466069}{10.1109/LRA.2024.3466069}.}
}

\begin{document}
\IEEEpubid{\begin{tabular}[t]{@{}c@{}}© 2024 IEEE.  Personal use of this material is permitted.  Permission from IEEE must be obtained for all other uses, in any current\\or future media, including reprinting/republishing this material for advertising or promotional purposes, creating new collective works,\\for resale or redistribution to servers or lists, or reuse of any copyrighted component of this work in other works.\end{tabular}}

\maketitle

\begin{abstract}
Model predictive control (MPC) has established itself as the primary methodology for constrained control, enabling general-purpose robot autonomy in diverse real-world scenarios.
However, for most problems of interest, MPC relies on the recursive solution of highly non-convex trajectory optimization problems, leading to high computational complexity and strong dependency on initialization.
In this work, we present a unified framework to combine the main strengths of optimization-based and learning-based methods for MPC.
Our approach entails embedding high-capacity, transformer-based neural network models within the optimization process for trajectory generation, whereby the transformer provides a near-optimal initial guess, or target plan, to a non-convex optimization problem.
Our experiments, performed in simulation and the real world onboard a free flyer platform, demonstrate the capabilities of our framework to improve MPC convergence and runtime.
Compared to purely optimization-based approaches, results show that our approach can improve trajectory generation performance by up to 75\%, reduce the number of solver iterations by up to 45\%, and improve overall MPC runtime by 7x without loss in performance.

\noindent Project website and code: \href{https://transformermpc.github.io}{https://transformermpc.github.io}
\end{abstract}

\IEEEpeerreviewmaketitle

\begin{figure*}[t]
\centering
    \includegraphics[width=0.87\textwidth]{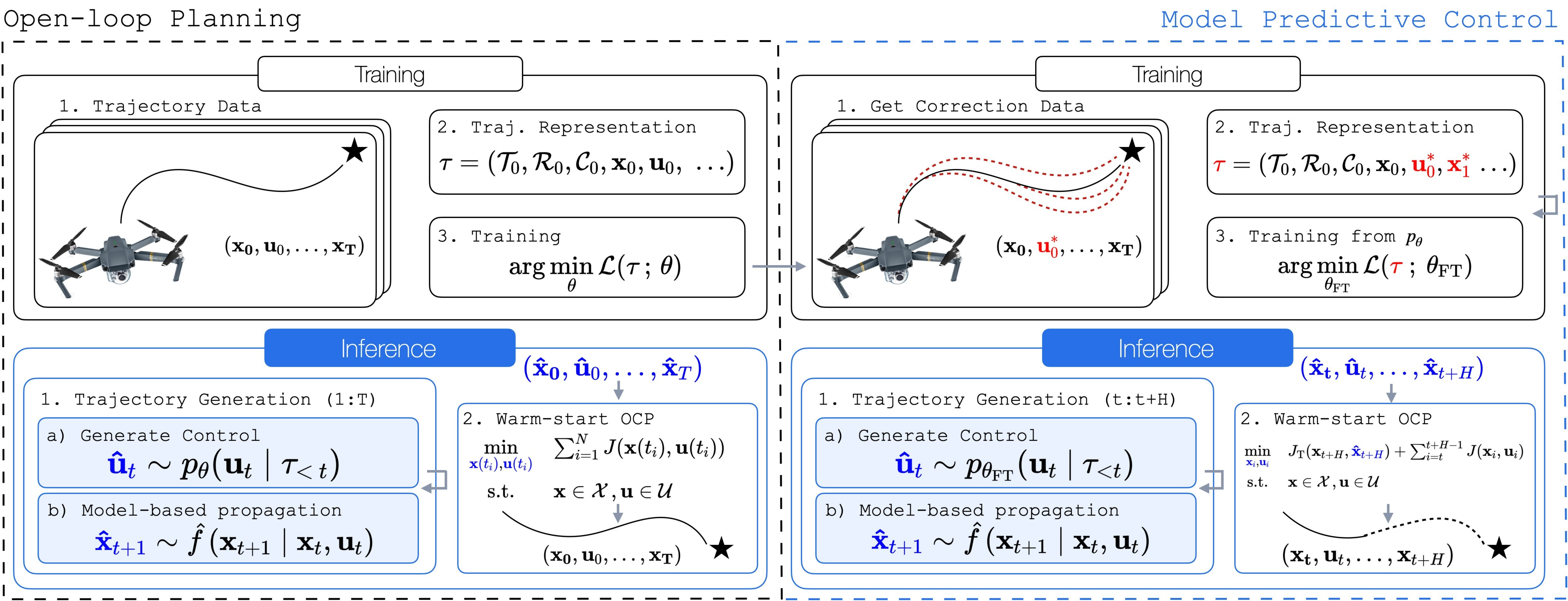}
    \caption{We propose a framework to combine high-capacity sequence models and optimization for MPC. The core idea is to train a transformer model to generate near-optimal trajectories (top), which can be used to warm-start optimal control problems (OCPs) at inference (bottom). We leverage methods introduced in~\cite{GuffantiGammelliEtAl2024} as a pre-training step, whereby a transformer is trained on pre-collected (open-loop) trajectory data (top left) to provide an initial guess for the full OCP (bottom left). \black{For effective MPC, we fine-tune the model (top right) through closed-loop corrections (red) and use it to provide both an initial guess for the OCP and a target state to approximate future cost within the short-sighted problem (bottom right).}}
    \label{fig:fig1}
\end{figure*}

\section{Introduction}
\IEEEPARstart{T}{rajectory} generation is crucial to achieving reliable robot autonomy, endowing autonomous systems with the capability to compute a state and control trajectory that simultaneously satisfies constraints and optimizes mission objectives.
As a result, trajectory generation problems have been formulated in many practical areas, including space and aerial
vehicles
~\cite{MalyutaEtAl2021, MellingerEtAl2011},
robot motion planning~\cite{MohananlEtAl2018},
chemical processes~\cite{MorariEtAl1999}, and more.
Crucially, the ability to solve the trajectory generation problem in real time is pivotal to safely operate within real-world scenarios, allowing the autonomous system to rapidly recompute an optimal plan based on the most recent information.

Motivated by its widespread applications, a collection of highly effective solution strategies exist for the trajectory generation problem. 
For example, numerical optimization provides a systematic mathematical framework to specify mission objectives as costs or rewards and enforce state and control specifications via constraints~\cite{BettslEtAl1998}.
However, for most problems of interest, the trajectory optimization problem is almost always nonconvex, leading to high computational complexity, strong dependency on initialization, and a lack of guarantees of either obtaining a solution or certifying that a solution does not exist~\cite{BoydEtAl2009}.
The above limitations are further exacerbated in the case of model predictive control (MPC) formulations, where the trajectory generation problem needs to be solved repeatedly and in real-time as the mission evolves, enforcing strong requirements on computation time. 
Moreover, MPC formulations typically require significant manual trial-and-error in defining ad-hoc terminal constraints and cost terms to, e.g., achieve recursive feasibility or exhibit short-horizon behavior that aligns with the full trajectory generation problem.
\IEEEpubidadjcol

Beyond methods based on numerical optimization, recent advances in machine learning (ML) have motivated the application of learning-based methods to the trajectory generation problem~\cite{LevineFinnEtAl2016}. 
ML approaches are typically highly computationally efficient and can be optimized for (potentially nonconvex) performance metrics from high-dimensional data (e.g., images).
However, learning-based methods are often sensitive to distribution shifts in unpredictable ways, whereas optimization-based approaches are more readily characterized both in terms of robustness and out-of-distribution behavior.
Additionally, ML methods often perform worse on these problems due to their high-dimensional action space, which increases variance in, e.g., policy-gradient algorithms~\cite{WuEtAl2018, ZhangEtAl2021}.
As a result, real-world deployment of learning-based methods has so far been limited within safety-critical applications.

In this work, we propose a framework to exploit the specific strengths of optimization-based and learning-based methods for trajectory generation, specifically tailored for MPC formulations (\cref{fig:fig1}).
By extending the framework introduced in~\cite{GuffantiGammelliEtAl2024}, we propose a pre-train-plus-fine-tuning strategy to train a transformer to generate near-optimal state and control sequences and show how this allows to (i) warm-start the optimization with a near-optimal initial guess, leading to improved performance and faster convergence, and (ii) provide \textit{long-horizon} guidance to \textit{short-horizon} problems in MPC formulations, avoiding the need for expensive tuning of cost terms or constraints within the optimization process.
Crucially, we show how the proposed fine-tuning scheme results in substantially improved robustness to distribution shifts caused by closed-loop execution and how the injection of learning-based guidance within MPC formulations drastically decreases the loss in performance due to the reduction of the planning horizon, enabling the solution of substantially smaller optimization problems without sacrificing performance.

The contributions of this paper are threefold:
\begin{itemize}
    \item We present a framework to combine the strengths of \textit{offline} learning and \textit{online} optimization for efficient trajectory generation within \black{MPC formulations}.
    \item \black{We investigate design and learning strategies within our framework, assessing the impact of fine-tuning on MPC execution and the benefits of learned terminal cost definitions to mitigate the inherent myopia in \textit{short-horizon} MPC formulations.}
    \item \black{Through experiments performed in simulation 
    (i.e., spacecraft rendezvous and quadrotor control)
    and real-world robotic platforms (i.e., a free flyer testbed), whereby we assume perfect state and scene estimation,} we show how our framework substantially improves the performance of off-the-shelf trajectory optimization methods in terms of cost, runtime, and convergence rates, demonstrating its applicability within real-world scenarios characterized by nonlinear dynamics and constraints.
\end{itemize}

\section{Related Work}
\label{sec:related_work}
Our work is closely related to previous approaches that
combine learning and optimization for control~\cite{IchterHarrisonEtAl2018, BansalEtAl2020, LewSinghEtAl2023},
exploit high-capacity neural networks
for control~\cite{ChenEtAl2021, JannerEtAl2021, ChiEtAl2023}, and methods for warm-starting nonlinear optimization problems~\cite{BanerjeeEtAl2020, CauligiEtAl2020, ChenWangEtAl2022, GuffantiGammelliEtAl2024}, 
providing a way to train general-purpose trajectory generation models that
guide the solution of inner optimization problems.

Numerous strategies have been developed for learning to control via formulations that leverage optimization-based planning as an inner component.
For instance, prior work focuses on developing hierarchical formulations, where a high-level (learning-based) module generates a waypoint-like representation for a low-level planner, e.g., based on sampling-based search~\cite{IchterHarrisonEtAl2018},
trajectory optimization~\cite{LewSinghEtAl2023}, or model-based planning~\cite{BansalEtAl2020}.
Within this context, the learning-based component is typically trained through either (online) reinforcement learning or imitation of oracle guidance algorithms.
An alternate strategy consists of directly optimizing through an inner controller. A large body of work has focused on exploiting exact solutions to the gradient of (convex) optimization problems at fixed points~\cite{AmosJimenezEtAl2018, AgrawalAmosEtAl2019}, allowing the inner optimization to be used as a generic component in a differentiable computation graph (e.g., a neural network).
\black{
On the contrary, our approach targets non-convex problems and leverages the output of a learning-based module as input to the unmodified trajectory optimization problems. This simplifies non-convex trajectory planning, avoiding the potential misalignment caused by intermediate problem representations that do not necessarily correlate with the downstream task (e.g., fixed-point iteration in value function learning)}.

Our method is closely related to recent work that leverages high-capacity generative models for control.
For example, prior work has shown how transformers~\cite{ChenEtAl2021, JannerEtAl2021} and diffusion models~\cite{ChiEtAl2023} trained via supervised learning on pre-collected trajectory data are amenable to both model-free feedback control~\cite{ChenEtAl2021} or (discrete) model-based planning~\cite{JannerEtAl2021}. These methods have two main drawbacks: (i) they typically ignore non-trivial state-dependent constraints, a setting that is both extremely common in practice and challenging for purely learning-based methods and (ii) they do not exploit all the information available to system designers, which typically includes approximate knowledge of the system dynamics.
As in~\cite{GuffantiGammelliEtAl2024}, our method alleviates both of these shortcomings by (i) leveraging online trajectory optimization to enforce non-trivial constraint satisfaction, and (ii) taking a model-based view to transformer-based trajectory generation and leveraging readily available approximations of the system dynamics to improve autoregressive generation.

Lastly, prior work has explored the idea of using ML to warm-start the solution of optimization problems.
While the concept of warm-starting optimization solvers is appealing in principle, current approaches are typically (i) limited by the choice of representation for the output of the ML model (e.g., a fixed degree polynomial)~\cite{BanerjeeEtAl2020, CauligiEtAl2020},
(ii) restricted to the open-loop planning setting~\cite{GuffantiGammelliEtAl2024}, whereby the impact of distribution shifts due to closed-loop execution~\cite{RajaramanEtAl2020} (e.g., model mismatch, stochasticity, etc.) is often overlooked\black{, and (iii) confined to linear constraints~\cite{ChenWangEtAl2022}.}
To address these limitations, we present an extended and revised version of~\cite{GuffantiGammelliEtAl2024}, whereby we \black{(i) build on prior work and use transformer network for effective trajectory modeling, (ii) introduce a pre-train-plus-fine-tuning learning strategy to maximize closed-loop performance, (iii) develop an MPC formulation that enables the transformer to provide long-horizon guidance to short-horizon problems, and (iv) augment the trajectory representation and the transformer architecture to handle conditioning on a target state.}
Crucially, we take full advantage of the transformer's autoregressive generation capabilities within receding horizon control formulations, enabling the same model to be used with varying planning horizon specifications.

\section{Problem Statement}
\label{sec:problem}
Let us consider the time-discrete optimal control problem (OCP):
{\fontsize{9pt}{11.5pt} \selectfont
\begin{mini!}|l|[2]
    {\bx(t_i),\bu(t_i)}
    {\calJ = \sum_{i=1}^{N}{\cost\left(\bx(t_i), \bu(t_i)\right)}\label{opt:cost}}{\label{optimizationProblem}}{}
    \addConstraint{\bx(t_{i+1}) = \boldf\left(\bx(t_i), \bu(t_i)\right) \hspace{5mm} \forall i \in \left[1,N\right]}{\label{constr:dynamics}}
    \addConstraint{\bx(t_i) \in \calX_{t_i}, \, \bu(t_i) \in \calU_{t_i} \hspace{7.3mm} \forall i \in \left[1,N\right],}{\label{constr:equality}}
\end{mini!}}where $\bx(t_i) \in \real^{\statedim}$ and $\bu(t_i) \in \real^{\controldim}$ are respectively the $\statedim$-dimensional state and $\controldim$-dimensional control vectors,
$\cost : \real^{\statedim + \controldim} \rightarrow \real$ defines the running cost, $\boldf:\real^{\statedim+\controldim} \rightarrow \real^{\statedim}$ represents the system dynamics, $\calX_{t_i}$ and $\calU_{t_i}$ are generic state and control constraint sets, and $\horizon \in \integer$ defines the number of discrete time instants $t_i$ over the full OCP horizon $T$.

In particular, we consider a receding-horizon reformulation of Problem (\ref{optimizationProblem}), whereby we recursively solve an OCP characterized by a moving (and typically shorter) horizon $\calH = \left[h, \rhchorizon\right] \subseteq \left[1, N\right]$, with $h$ being the moving initial timestep and $H \in \left(0,N\right]$ the length of the
horizon. 
To ensure desirable long-horizon behavior from the solution of a series of short-horizon problems, MPC formulations typically require the definition of a terminal cost $\terminalcost$ or constraint set $\calX_{\rhchorizon}$.
Specifically, we will consider cost functions of the form:
{\fontsize{9pt}{11.5pt} \selectfont
\begin{equation}
    \label{ocp:closed_loop_cost}
    \calJ = \terminalcost \left( \bx(t_{\rhchorizon}) \right) + \sum_{i=h}^{\rhchorizon-1} \cost \left( \bx(t_i), \bu(t_i) \right),
\end{equation}}
where the running cost $\cost$ is evaluated exclusively over $\calH$ and $\terminalcost : \real^{\statedim} \rightarrow \real$ is the terminal cost function.

In this work, we explore approaches to address Problem (\ref{ocp:closed_loop_cost}) by incorporating high-capacity neural network models within the optimization process for trajectory generation, explicitly tailored for closed-loop performance.


\section{Trajectory Optimization via Sequence Modeling}
\label{sec:method}
We consider a strategy for trajectory generation whereby state and control sequences $\bX\! =\! \left( \bx_1, ..., \bx_N\! \right)\!,\bU\! =\! ( \bu_1, ..., \bu_N\! )$ are obtained through the composition of two components, 
{\fontsize{9pt}{11.5pt} \selectfont
\begin{align}
\left( \hat \bX, \hat \bU \right) & \sim p_{\theta}(\bX, \bU \, | \, \initcond) \label{eq:cond_prob}\\
\left( \bX, \bU \right) & = \opt \left(\bx(t_1), \hat \bX, \hat \bU \right),
\end{align}
}
where $p_{\params}(\cdot)$ denotes the conditional probability distribution over trajectories (given an initial condition $\initcond$) learned by a transformer model with parameters $\params$, $\opt$ denotes the trajectory optimization problem, and $\hat \bX$, $\hat \bU$ denote the predicted (state and control) trajectories that are taken as an input to the optimization problem.
The initial condition $\initcond$ is used to inform trajectory generation in the form of, e.g., an initial state $\bx(t_1)$, a target state $\bx(t_N)$ to be reached by the end of the trajectory, or other performance-related parameters.
In this section, we will first dive into the details of the trajectory representation for sequence modeling. 
We will then discuss both open-loop pre-training and MPC fine-tuning formulations of our approach, together with specific inference algorithms for open-loop planning and MPC.

\subsection{Trajectory Representation}
At the core of our approach is the treatment of trajectory data as a sequence to be modeled by a transformer model~\cite{GuffantiGammelliEtAl2024}.
Specifically, given a pre-collected dataset of trajectories of the form
    $\traj_{\mathrm{raw}} = \left(\bx_1, \bu_1, r_1, \dots, \bx_N, \bu_N, r_N \right),$
where $\bx_i$ and $\bu_i$ denote the state and control at time $t_i$, and $r_i = - \cost(\bx(t_i), \bu(t_i))$ is the instantaneous reward, or negative cost, we define the following trajectory representation:
{\fontsize{9pt}{11.5pt} \selectfont \begin{equation}
    \label{eq:trajectory:representation}
    \traj = \left(\calT_1, \calR_1, \calC_1, \bx_1, \bu_1, \dots,\calT_N, \calR_N, \calC_N, \bx_N, \bu_N \right),
\end{equation}}
where $\calT_i \in \real^{\statedim}, \,\,  \calR_i \in \real$ and $ \calC_i \in \integer^+$ represent a set of performance parameters that allow for effective trajectory generation. 
In particular, we define $\calT_i$ as the \textit{target state}, i.e., the state we wish to reach by the end of the trajectory, that could be either constant during the entire trajectory or time-dependent. 
We further define $\calR_i$ and $\calC_i$ as the \textit{reward-to-go} and \textit{constraint-violation-budget} evaluated at $t_i$, respectively.
Formally, as in~\cite{GuffantiGammelliEtAl2024}, we define $\calR_i$ and $\calC_i$ to express future optimality and feasibility of the trajectory as:
{\fontsize{9pt}{11.5pt} \selectfont
\begin{equation}
    \label{eq:ctg}
    \calR(t_i)\! =\! \sum_{j = i}^{\horizon}\!  r_j,\! \quad C(t_i)\! =\! \sum_{j = i}^{N}\!{{\textsf{C}}_j},\! \quad  {\textsf{C}}_j\! =\! 
\begin{cases}
    1,\! & \text{if }\! \exists \bx_j,\! \bu_j\! \notin\! \calX_{j}\!, \calU_{j}\! \\
    0,\! & \text{otherwise}.
\end{cases}
\end{equation}}
This definition of the performance parameters is appealing for two reasons: (i) during training, the performance parameters are easily derivable from raw trajectory data by applying (\ref{eq:ctg})
for $\calR_i$ and $\calC_i$, and by setting $\calT_i = \bx_{\horizon}$, (ii) at inference, according to (\ref{eq:cond_prob}), it allows to condition the generation of predicted state and control trajectories $\hat \bX$ and $\hat \bU$ through user-defined initial conditions $\initcond$. 
Namely, given user-defined $\initcond = (\calT_1, \calR_1, \calC_1, \bx_1)$, a successfully trained transformer should be able to generate a trajectory $(\hat \bX, \hat \bU)$ starting from $\bx(t_1)$ that achieves a reward of $\calR_1$, a constraint violation of $\calC_1$, and terminates in state $\calT_1$.

\subsection{Open-loop Pre-training}
\label{sec:open_loop_ws_traj_gen}
In what follows, we introduce the pre-training strategy used to obtain a transformer model capable of generating near-optimal trajectories $\hat{\bX} = \left( \hat{\bx}_1, ..., \hat{\bx}_N \right),\hat{\bU} = \left( \hat{\bu}_1, ..., \hat{\bu}_N \right)$ in an open-loop setting.

\noindent\textbf{Dataset generation.}
The first step entails generating a dataset for effective transformer training. 
To do so, we generate $N_D$ trajectories by repeatedly solving Problem (\ref{optimizationProblem}) with randomized initial conditions and target states, and then re-arranging the raw trajectories according to (\ref{eq:trajectory:representation}).
We construct the datasets to include both solutions to Problem (\ref{optimizationProblem}) as well as to its relaxations. 
We observe that diversity in the available trajectories is crucial to enable the transformer to learn the effect of the performance metrics; for example, solutions to relaxations of Problem (\ref{optimizationProblem}) will typically yield trajectories with low cost and non-zero constraint violation (i.e., high $\calR$ and $\calC > 0$), while direct solutions to Problem (\ref{optimizationProblem}) will be characterized by higher cost and zero constraint violation ( i.e., lower $\calR$ and $\calC = 0$), allowing the transformer to learn a broad range of behaviors from the specification of the performance parameters.

\noindent\textbf{Training.}
We train the transformer with the standard teacher-forcing procedure used to train sequence models.
Specifically,
denoting the L2-norm as $|| \cdot ||_2$, we optimize the following mean-squared-error (MSE) loss function:
{\fontsize{9pt}{11.5pt} \selectfont
\begin{equation}
    \label{eq:ol_loss_function}
    \olloss(\tau) = \sum_{n=1}^{N_D} \sum_{i=1}^{N} \left( \lVert \bx^{(n)}_i - \hat{\bx}^{(n)}_i \rVert_2^2 + \lVert \bu^{(n)}_i - \hat{\bu}^{(n)}_i \rVert_2^2 \right),
\end{equation}}
where $n$ denotes the $n$-th trajectory sample from the dataset, $\hat{\bx}^{(n)}_i\!\sim\!p_{\params}( \bx^{(n)}_i\!\mid\!\tau^{(n)}_{<t_i}), \hat{\bu}^{(n)}_i\!\sim\!p_{\params}(\bu^{(n)}_i\!\mid\!\tau^{(n)}_{<t_i}, \bx^{(n)}_i)$ are the one-step predictions for both state and control vectors, and where we use $\tau_{<t_i}$ to denote a trajectory spanning timesteps $t \in \left[t_1, t_{i-1}\right]$.

\noindent\textbf{Inference.}
Once trained, the transformer can be used to autoregressively generate an open-loop trajectory, i.e., $(\hat{\bX},\hat{\bU}) \sim p_{\params}(\bX, \bU \, | \, \initcond)$, from a given initial condition $\initcond = (\calT_1, \calR_1, \calC_1, \bx_1)$ that encodes the initial state, target state, and desired performance metrics.
Given $\initcond$, the autoregressive generation process is defined as follows:
(i) the transformer generates a control $\bu_1$, (ii) \black{building on the findings of~\cite{GuffantiGammelliEtAl2024}, we take a model-based view on autoregressive generation, whereby the next state $\bx_2$ is generated according to a (known) approximate dynamics model $\dynapprox(\bx, \bu)$---a reasonable assumption across the problem settings we investigate---}, (iii) the performance metrics are updated by decreasing the reward-to-go $\calR_1$ and constraint-violation-budget $\calC_1$ by the instantaneous reward $r_1$ and constraint violation ${\textsf{C}}(t_1)$, respectively, (iv) the target state $\calT_1$ is either kept constant or updated to a new target state, and (v) the previous four steps are repeated until the horizon is reached.
To specify the performance parameters, we select $\calR_1$ as a (negative) quantifiable lower bound of the optimal cost and $\calC_1 = 0$, incentivizing the generation of near-optimal and feasible trajectories.

\subsection{Long-horizon Guidance for Model Predictive Control}
The core of our approach relies on leveraging the recursive computation within sequence models for effective model predictive control.
Specifically, given a planning horizon $H$, we fine-tune a transformer to generate trajectories $\hat \bX_{h:\rhchorizon} = (\hat \bx_{h}, \dots, \hat \bx_{\rhchorizon}), \, \hat \bU_{h:\rhchorizon} = (\hat \bu_{h}, \dots, \hat \bu_{\rhchorizon})$.
In our formulation, the generated trajectories are used to provide a warm-start to the trajectory optimization problem, as in the open-loop planning scenario, but also, crucially, to specify an effective terminal cost $\terminalcost$.
As a result, we use the transformer to (i) improve runtime by enabling the solution of smaller optimization problems and (ii) define learned terminal cost terms and avoid expensive cost-tuning strategies.

\noindent\textbf{Closed-loop fine-tuning.} 
Rather than naively applying the transformer trained on open-loop data to the MPC setting, we devise a fine-tuning scheme to achieve effective closed-loop behavior. 
It is well-known that methods based on imitation learning are exposed to severe error compounding when applied in closed-loop~\cite{RajaramanEtAl2020}.
The main reason behind this error compounding is the covariate shift problem, i.e., the fact that the \textit{actual} performance of the learned policy depends on its own state distribution, whereas \textit{training} performance is only affected by the expert’s state distribution (i.e., the policy used to generate the data).
To address this problem, we resort to iterative algorithms, namely $\dagname$~\cite{RossEtAl2011}, in which the current policy interacts with the training environment to collect trajectory data and define a new dataset for training based on expert corrections.
In our framework, we fine-tune the transformer in simulation \black{according to the same approximate model $\dynapprox (\bx, \bu)$ used for the pre-training,} hence the state distribution shift arises uniquely from the differences between open-loop and closed-loop cost functions---see (\ref{opt:cost}) vs (\ref{ocp:closed_loop_cost}).
Our tailored $\dagname$ algorithm is defined according to three main design choices (Alg. \ref{alg:DAGGER}).
First, to incentivize the transformer to learn optimal solutions for the full-horizon OCP, we define the expert policy as the solution to the full horizon Problem (\ref{optimizationProblem}).
Second, to promote successful generalization across various planning horizons $H$, we randomly sample the exploration policy parameter $H$ from a discrete set that uniformly spans short to full-horizon problem formulations.
Third, as is usual for $\dagname$ implementations, we alternate between two exploration policies: the transformer from the last dagger iteration and the expert policy.
\black{Note that
using $\dagname$ from scratch would decrease the efficiency in the exploration and increase
the risk of getting stuck in locally optimal policies.}
\SetAlFnt{\small}
\begin{algorithm}[t!]
    \caption{Fine-tuning strategy for MPC}
    \SetAlgoVlined
    \label{alg:DAGGER}
    \KwIn{\black{Pre-trained $p_{\params_0}$, Expert $\pi^*$, Open-loop data $\calD_{\ol}$}}
    
    Initialize $\calD_{\ft} \leftarrow \{\calD_{\ol}\}$\;
    \For{$i=0$  to \fontfamily{qcr}\selectfont dagger iterations}{
        \For{\fontfamily{qcr}\selectfont num trajectories}{
            \black{Select exploration policy $\exppol^t = \{p_{\params_i}, \pi^*\}, \, \forall t$}\;
            Sample planning horizon $H \in (0, N]$\;
            Get dataset $\calD_i = \{\bx, \pi^*(\bx)\}$ of states visited by $\exppol$ and controls given by $\pi^*$\;
        }
       Aggregate dataset $\calD_{\ft} \leftarrow \calD_{\ft} \cup \calD_{i}$\;
       \black{Update model $p_{\params_{i+1}} \leftarrow \mathrm{Train}\left( p_{\params_i},\calD_{\ft}\right)$}\;
    }
\end{algorithm}

\noindent\textbf{Inference.}
After fine-tuning, we use the transformer to generate a trajectory $(\hat \bX_{h:\rhchorizon}, \hat \bU_{h:\rhchorizon}) \sim p_{\params}(\bX_{h:\rhchorizon},  \bU_{h:\rhchorizon} \mid \tau_{<h})$ following the same five-step process defined in the open-loop scenario, whereby the autoregressive generation is executed for $H$ steps, rather than for the full OCP horizon $N$. 
We further use the generated trajectory to (i) warm-start the short-horizon OCP, and (ii) define a cost function of the form:
{\fontsize{9pt}{11.5pt} \selectfont
\begin{equation}
    \label{ocp:closed_loop_cost2}
    \calJ = \terminalcost \left( \bx(t_{\rhchorizon}), \hat \bx(t_{\rhchorizon}) \right) + \sum_{i=h}^{\rhchorizon} \cost \left( \bx(t_i), \bu(t_i) \right),
\end{equation}}
where the
function $\terminalcost(\bx_{\rhchorizon}, \hat \bx_{\rhchorizon})$ is defined as a distance metric that penalizes deviation from the predicted terminal state $\hat \bx_{\rhchorizon}$.
As a consequence of this formulation, we use the transformer---trained to generate near-optimal trajectories for the \textit{full} OCP---to incentivize solutions that better align with the long-horizon problem.
\begin{figure*}[t]
\centering
    \includegraphics[width=0.87\linewidth]{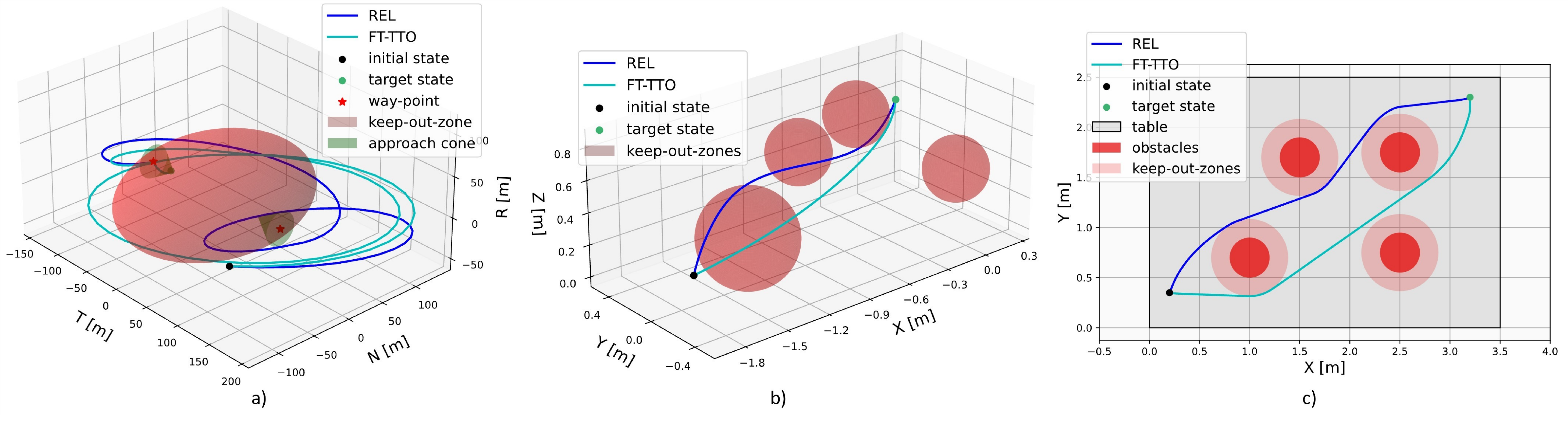}
    \caption{Visualization of the three
    scenarios considered in this work. For the tasks of (a) spacecraft rendezvous, (b) quadrotor and (c) free flyer control, we show three example trajectories obtained by warm-starting the SCP through a relaxation of the full OCP (blue) or through our proposed approach (cyan). Keep-out-zones and obstacles are denoted by the red shaded areas.}
    \label{fig:scenarios}
\end{figure*}
\section{Experiments}
\label{sec:simulations}
In this section, we demonstrate the performance of our framework on three trajectory optimization problems: two in simulation (i.e., \textit{spacecraft rendezvous} and \textit{quadrotor control} scenarios) and one real-world robotic platform (i.e., a \textit{free flyer} testbed).
We consider three instantiations of Problem (\ref{optimizationProblem}) based on the following general formulation:
{\fontsize{9pt}{11.5pt} \selectfont
\begin{mini!}|l|[2]
    {\bx_i,\bu_i}
    {
    \sum_{i=1}^{N}{\lVert \bu_i \rVert_p^q}\label{opt2:cost}
    }{\label{optimizationProblem2}}{}
    \addConstraint{\black{\bx_{i+1} = \dynapprox\left(\bx_i, \bu_i\right) \hspace{17mm} \forall i \in \left[1,N\right]}}{\label{constr2:dynamics}}
    \addConstraint{\Gamma \left( \bx_i \right) \ge \mathbf{0}  \hspace{25mm} \forall i \in \left[1,N\right]}
    {\label{constr2:koz}}
    \addConstraint{
    \bx_1=\bx_{\start}, \quad \bx_{N+1}=\bx_{\goal}.
    }{\label{constr2:initial&final}}
\end{mini!}}
where the objective function in (\ref{opt2:cost}) expresses the minimization of the control effort (with $p, q$ denoting application-dependent parameters), \black{
(\ref{constr2:dynamics}) represents the (potentially nonlinear) approximate system dynamics,} (\ref{constr2:koz}) defines the obstacle avoidance constraints through a nonlinear distance function $\Gamma\!:\!\real^{m \times n_{x}}\! \rightarrow\! \real^{m}$ with respect to $m \in [1, M]$ non-convex keep-out-zones, \black{and (\ref{constr2:initial&final})
represents the initial and terminal state constraints.}

\noindent\textbf{Experimental design.}
While the specific formulations of the trajectory optimization problem will necessarily depend on the individual application, we design our experimental setup to follow some common desiderata.

First, we care to isolate the benefits of (i) the initial guess (in both open-loop and MPC settings) and (ii) learning the terminal cost for MPC; thus, we keep both the formulation and the solution algorithm for the OCP fixed and only evaluate the effect of different initializations.
In particular, we resort to sequential convex programming (SCP) methods for the solution of the trajectory optimization problem and compare different approaches that provide an initial guess for the sequential optimizer (or, for MPC, also a specification of the terminal cost).
Second, in the open-loop setting, we always compare our approach, identified as Transformers for Trajectory Optimization (TTO), with the following \textit{classes} of methods: \black{(i) a quantifiable cost \textit{Lower Bound} (LB), characterized by the solution of a relaxation (REL) to the full-horizon Problem (\ref{optimizationProblem2}) which ignores obstacle avoidance constraints from (\ref{constr2:koz})
and thus represents a (potentially infeasible) cost lower bound to the full problem and (ii) a state-of-the-art SCP approach where the initial guess to Problem (\ref{optimizationProblem2}) is provided by the solution to the REL relaxation described in (i)~\cite{BanerjeeEtAl2020,AlcanKyrki2022}.}
Third, in the model predictive control setting, we always consider terminal cost terms of the form $\terminalcost(\bx_{\rhchorizon}, \bar \bx) = \lVert \bx_{\rhchorizon} - \bar \bx \rVert_{P}$, where $\lVert \cdot \rVert_P$ denotes the weighted norm with respect to a diagonal matrix $P$ and compare different choices for $\bar \bx$ and $P$.
\black{Lastly, as introduced in Sec. \ref{sec:open_loop_ws_traj_gen}, to obtain diverse trajectories for offline training
we solve $N_d=400,000$ instances of the trajectory optimization problem, whereby $\frac{N_d}{2}$ are solutions to the full non-convex Problem (\ref{optimizationProblem2}), and $\frac{N_d}{2}$ to its REL relaxation.}

In our experiments, we care about three main performance metrics: cost, speed of optimization (expressed in terms of the number of SCP iterations),
and overall runtime. 
\black{As in~\cite{GuffantiGammelliEtAl2024}, throughout our experiments, 
we assume knowledge of the fixed operational environment and perfect state estimation at all times}. We further assume that the target state is reachable within $N$ discrete time instants.

\noindent\textbf{Transformer architecture.}
Our model is a transformer architecture designed to account for continuous inputs and outputs. 
Given an input sequence and a pre-defined maximum context length $K$, our model takes as input the last $5K$ sequence elements, one for each modality: target state, reward-to-go, constraint-violation-budget, state, and control.
The sequence elements are projected through a modality-specific linear transformation, obtaining a sequence of $5K$ embeddings.
As in~\cite{JannerEtAl2021}, we further encode an embedding for each timestep in the sequence and add it to each element embedding.
The resulting sequence of embeddings is processed by a causal GPT model consisting of six layers and six self-attention heads.
Lastly, the GPT architecture autoregressively generates a sequence of latent embeddings which are projected through modality-specific decoders to the predicted states and controls.

\subsection{Problem Description}
\label{sec:scenarios_description}
In this section, we describe our experimental scenarios, the definition of cost, state, and control variables, and the specific implementation of constraints in (\ref{constr2:dynamics}-\ref{constr2:initial&final}).

\noindent\textbf{Spacecraft rendezvous and proximity operations.}
We consider the task of performing an autonomous rendezvous, proximity operation, and docking (RPOD) transfer, in which a servicer spacecraft equipped with impulsive thrusters approaches and docks with a target spacecraft.
The relative motion of the servicer with respect to the target is described using the quasi-nonsingular Relative Orbital Elements (ROE) formulation, i.e. $\bx := \delta \boe \in \real^6$, \cite{KoenigGuffantiEtAl2017,GuffantiGammelliEtAl2024}. 
Cartesian coordinates in the Radial Tangential Normal (RTN) frame can be obtained through a time-varying linear mapping \cite{GuffantiGammelliEtAl2024} and employed for imposing geometric constraints. 

Problem (\ref{optimizationProblem2}) is detailed as follows:
(i) in (\ref{opt2:cost}), we select $p=2$ and $q=1$, to account for impulsive thrusters aligned with the desired control input;
(ii) in (\ref{constr2:initial&final}), $\bx_{\start}$ is defined by a random passively-safe relative orbit with respect to the target,
while $\bx_{\goal}$ is selected among two possible docking ports located on the $\mathrm{T}$ axis, as shown in Fig. \ref{fig:scenarios}a; 
(iii) in (\ref{constr2:dynamics}), $\delta \boe_{i+1} = \Phi_i \left(\Delta t \right) \left( \delta \boe_i + B_i \bu_i \right)$ is used to enforce dynamics constraints, with $\Phi_i (\Delta t) \in \real^{6\times6}$ and $B_i \in \real^{6\times3}$ being the linear time-varying state transition and control input matrices \cite{GuffantiGammelliEtAl2024}, $\Delta t$ the time discretization and $\bu_i = \Delta v_i \in \real^3$ the impulsive delta-velocity applied by the servicer; 
(iv) in (\ref{constr2:koz}), we consider an elliptical keep-out-zone (KOZ) centered at the target, limiting the time index $i$ to $\left[1,N_{\textrm{wp}}\right]$. 
We further consider two additional domain-specific constraints: (v) a zero relative velocity pre-docking waypoint to be reached at $i=N_{\textrm{wp}}$;
(vi) a second-order cone constraint enforcing the service spacecraft to approach the docking port inside a cone with aperture angle $\gamma_{\textrm{cone}}$ for $N_{\textrm{wp}} \le i \le N$, with axis $\mathbf{n}_{\textrm{cone}} \in \real^3$ and vertex $\delta \mathbf{r}_{\textrm{cone}} \in \real^3$ determined accordingly to the selected docking port.

\noindent\textbf{Quadrotor control.}
We consider the task of autonomous quadrotor flight, in which a quadrotor robot has to reach a target state while navigating an obstacle field, Fig. \ref{fig:scenarios}b. 
We use a Cartesian reference frame $\mathrm{O}_{\mathrm{xyz}}$, with the $z$-axis pointing upward, to define $\bx\!:=\!\left(\br,\bv\right)\! \in \!\real^6, \bu\! :=\! \bT\! \in\! \real^3$, with $\br, \bv, \bT \in \real^3$ being the quadrotor's position, velocity and thrust vectors, respectively. 

We define the following elements of Problem (\ref{optimizationProblem2}): (i) in (\ref{opt2:cost}), we select $p=q=2$ to adhere to the continuous nature of the control input; 
(ii) in (\ref{constr2:initial&final}),
we sample $\bx_{\start}$ and $\bx_{\goal}$ from pre-determined start and goal regions;
(iii) in (\ref{constr2:dynamics}), system dynamics are given by the nonlinear model
$\br_{i+1} = \br_i + \bv_i \Delta t, \ \bv_{i+1} = \bv_i + \frac{1}{m} \left( -\beta_{\textrm{drag}} \lVert \bv_i \rVert_2^2 + \bu_i\right) \Delta t$,
with $m$ and $\beta_{\textrm{drag}}$ being the mass and the drag coefficient of the drone, respectively; 
(iv) in (\ref{constr2:koz}), we consider multiple spherical KOZs distributed between the start and goal regions.

\begin{figure}[t]
\centering
    \includegraphics[width=0.85\linewidth]{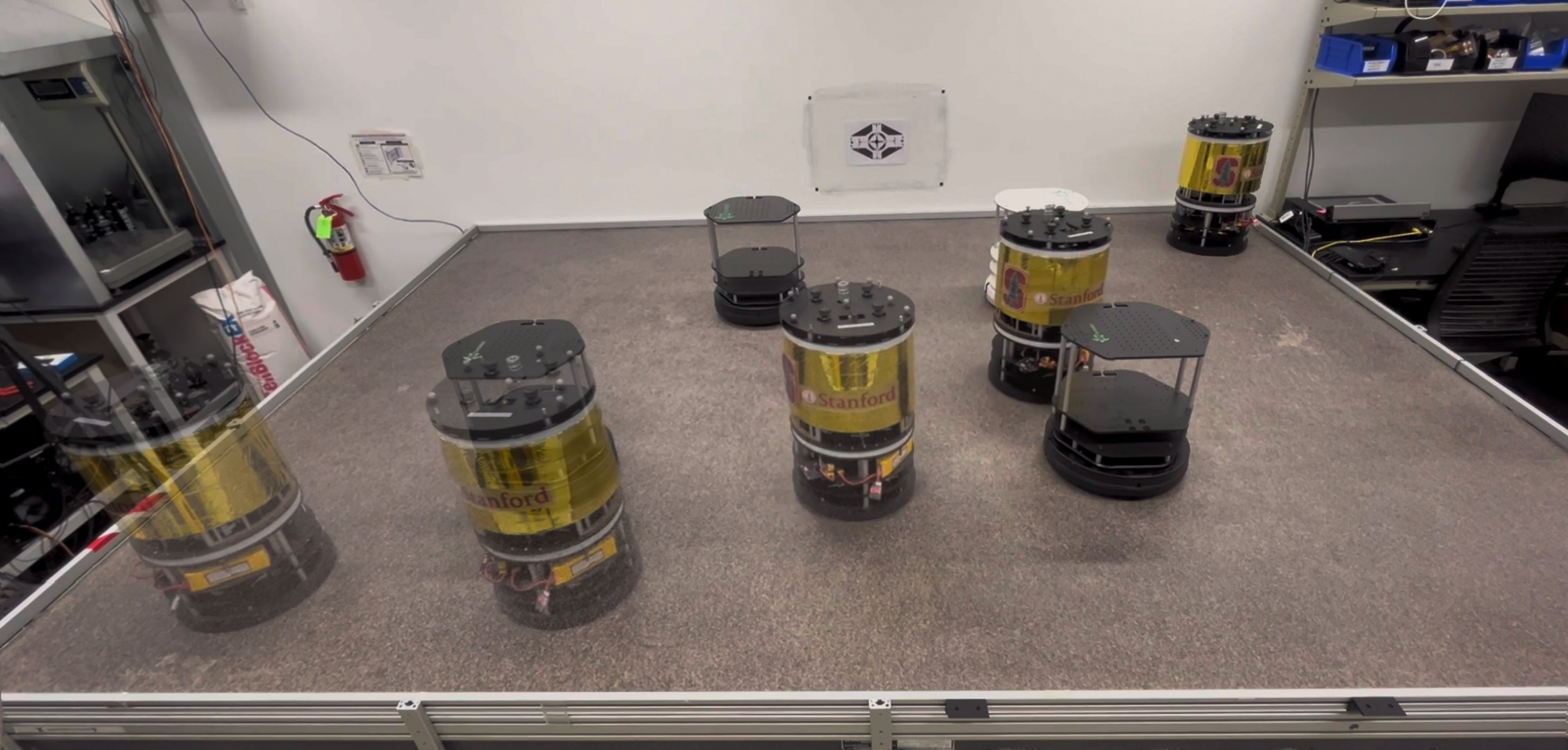}
    \caption{Free flyer testbed and real-world execution of the FT-TTO trajectory in \cref{fig:scenarios}c. Transparency identifies the time progression from the start (faded) to the end (opaque) of the trajectory.}
    \label{fig:testbed}
\end{figure}
\noindent\textbf{Free flyer testbed.}
Lastly, we consider a real-world free flyer platform in which a floating robot---the free flyer---can move over a granite table in absence of friction, simulating space flight. 
We consider the task of controlling the free flyer, equipped with eight ON-OFF thrusters allowing bi-dimensional roto-translational motion, to achieve a target state by traversing an obstacle field. 
The simulation scenario and the real-world testbed are reported in \cref{fig:scenarios}c and \ref{fig:testbed}, respectively.
We use a global Cartesian frame $\mathrm{O}_{\mathrm{xy}}$ to define $\bx:=\left(\br,\psi,\bv,\omega\right) \in \real^6, \bu:=R_{\mathrm{GB}} (\psi) \Lambda \Delta \bV \in \real^3$, where $\br,\bv \in \real^2$ and $\psi, \omega \in \real$ are respectively the position, velocity, heading angle and angular rate of the free flyer, $\Delta \bV \in \real^8$ consists of the impulsive delta-velocity applied by each thruster, $\Lambda \in \real^{3 \times 8}$ is the thrusters' configuration matrix and $R_{\mathrm{GB}} \in \real^{3 \times 3}$ is the rotation matrix from the body reference frame of the free flyer to $\mathrm{O}_{\mathrm{xy}}$.

Problem (\ref{optimizationProblem2}) is specified as follows: 
(i) in (\ref{opt2:cost}), we select $p=q=1$ as the ON-OFF thruster are not aligned with the control input;
(ii) in (\ref{constr2:initial&final}),
$\bx_{\start}$ and $\bx_{\goal}$ are sampled from pre-determined start and goal regions; 
(iii) in (\ref{constr2:dynamics}), the dynamics are propagated using the impulsive model $\bx_{i+1} = \bx_i + \mathrm{diag}\left([\Delta t, \Delta t, \Delta t, 1, 1, 1]\right) \bu_i$;
(iv) in (\ref{constr2:koz}), we consider multiple spherical KOZs distributed between the start and goal regions, comprising the obstacles' and the free flyer's radii. We further consider two domain-specific constraints: (v) \black{nonlinear actuation limits defined by $0 \le \Lambda^{-1}R_{\mathrm{GB}}^{-1}\bu \le \Delta \bV_{\mathrm{max}}$, with $\Delta \bV_{\mathrm{max}} = \frac{T \Delta t}{m}$, where $T$ and  $m$ denote the thrust level of the thrusters and the mass of the free flyer, respectively;} (vi) state bounds, i.e. $\bx \in \calX_{\mathrm{table}}$, with $\calX_{\mathrm{table}}$ defining the table region.

\subsection{Open-loop Planning}
As a first experiment, we aim to solve the trajectory generation problem in an open-loop setting. 
As shown in Fig. \ref{fig:scenarios}, this results in a single optimal trajectory that can be robustly tracked by a downstream control system.
Specifically, as a form of comparison, we warm-start the SCP using trajectories generated by (i) \black{solving the REL relaxation of Problem (\ref{optimizationProblem2})~\cite{BanerjeeEtAl2020,AlcanKyrki2022}}
(ii) the transformer \textit{before} fine-tuning (TTO), and (iii) the transformer \textit{after} fine-tuning (FT-TTO). 
To initialize the generation process, \black{we set the performance parameter $\calR_1$ to the negative cost of the REL solution and $\calC_1 = 0$.}
\black{For each scenario, we evaluate trajectory generation performance across 40,000 random problem specifications generated as in Sec. \ref{sec:scenarios_description} and ordered according to a \textit{non-convexity factor}} computed as $\frac{\calC_1^j}{\calC_{\textrm{max}}}$, where $\calC_1^j$ is the number of KOZ constraint violations observed in the REL solution to each problem specification $j \in [1, \, \, \text{40,000}]$ and $\calC_{\textrm{max}}\! = \!\max_j \calC_1^j$ is the maximum constraint violation observed in the scenario.
In practice, we use the non-convexity factor as a direct metric of the difficulty of the considered scenario, whereby scenarios with a factor of 0 represent OCPs for which there exists a convex solution and hence the solution of REL is optimal for the full problem;
whereas scenarios with a factor of 1 represent OCPs for which the closest convex relaxation is highly infeasible and where a better selection of the warm-start can have a larger impact on the solution to the non-convex problem.
\label{sec:open_loop_res}
\begin{figure}[t]
\centering
    \includegraphics[width=0.87\linewidth]{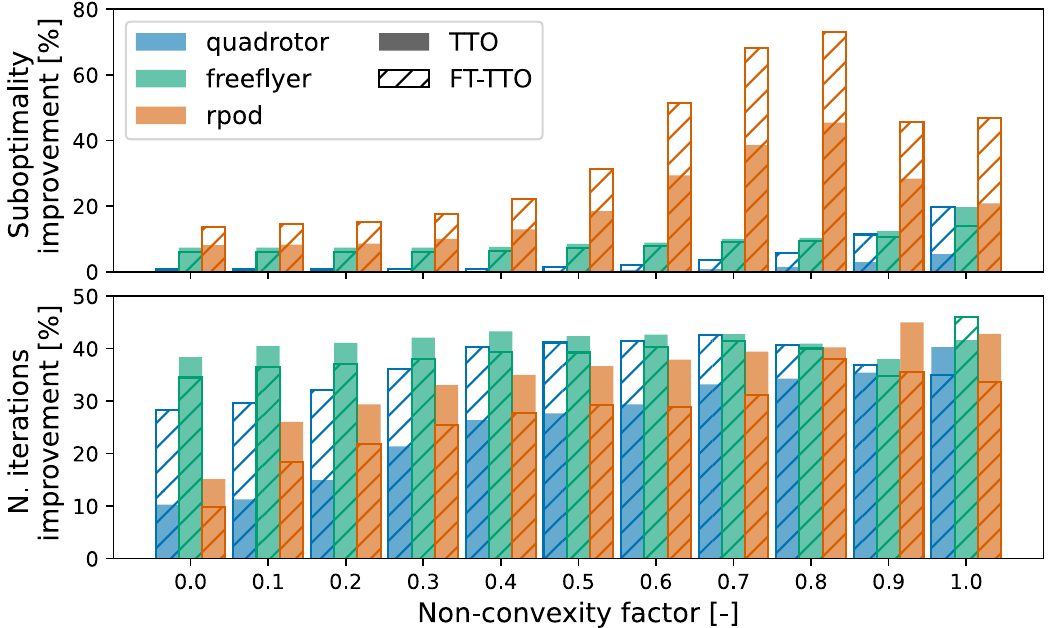}
    \caption{Percentage improvement in terms of cost suboptimality (top) and number of SCP iterations (bottom) with respect to REL achieved by warm-starting the SCP with FT-TTO and TTO. Each bar represents the improvement averaged over non-convexity factors greater or equal to the corresponding x-axis value.}
    \label{fig:open_loop_res}
\end{figure}

Results in \cref{fig:open_loop_res} confirm the findings of~\cite{GuffantiGammelliEtAl2024} and show a clear advantage in warm-starting the non-convex problem with TTO, which strongly outperforms initial guesses based on problem relaxations, both in terms of cost and number of SCP iterations.
Additionally, \cref{fig:open_loop_res} clearly highlights the benefits of the proposed fine-tuning strategy on open-loop performance. 
In particular, \cref{fig:open_loop_res} (top) shows how FT-TTO is able to reduce the sub-optimality gap with respect to the lower bound when compared to the model before fine-tuning: 
FT-TTO achieves approximately a 2x cost improvement compared to TTO in the case of spacecraft RPOD (peak value $75 \%$) and quadrotor control (peak value $20 \%$), while resulting in similar cost performance in the free flyer testbed simulation.
Furthermore, in all the scenarios considered, warm-starting with transformers consistently reduces the number of SCP iterations when compared to initial guesses based on REL.
\black{Specifically, \cref{fig:open_loop_res} (bottom) shows how both TTO and FT-TTO achieve at least a $10\%$ reduction in SCP iterations across all scenarios and non-convexity levels, achieving up to $40\%$ improvement for high values of the non-convexity factor.}
Crucially, the substantial cost reduction and the comparable results in SCP iterations demonstrate that, from an open-loop perspective, the fine-tuning of FT-TTO does not cause any degradation in performance, improving open-loop behavior.

\subsection{Model Predictive Control}
\label{sec:closed_loop_res}
The second part of our experiments focuses on solving the trajectory generation problem within the MPC scheme.
\black{We compare two state-of-the-art approaches with our transformer-based framework for the generation of the warm-start and the definition of the terminal cost $\terminalcost(\bx_{\rhchorizon}, \bar \bx) = \lVert \bx_{\rhchorizon} - \bar \bx \rVert_{2}$: (i) REL-MPC
solves the full-horizon REL problem and defines $\bar \bx$ as the state obtained by the REL solution at time $t_{\rhchorizon}$, i.e. $\bar \bx := \bx_{\rhchorizon}^{(REL)}$~\cite{AlcanKyrki2022}, (ii) DIST-MPC
defines $\terminalcost$ to encode the distance from the goal, i.e. $\bar \bx := \bx_{\textrm{goal}}$, and uses the REL solution to this short-horizon problem as warm-start~\cite{NguyenKamelEtAl2021}},
(iii) TTO-MPC and (iv) FT-TTO-MPC employ the transformer models before and after $\dagname$ fine-tuning, respectively, to generate a short-horizon warm-start and define $\bar \bx := \hat{\bx}_{\rhchorizon}$.

\begin{figure}[t!]
\centering
    \includegraphics[width=0.86\linewidth]{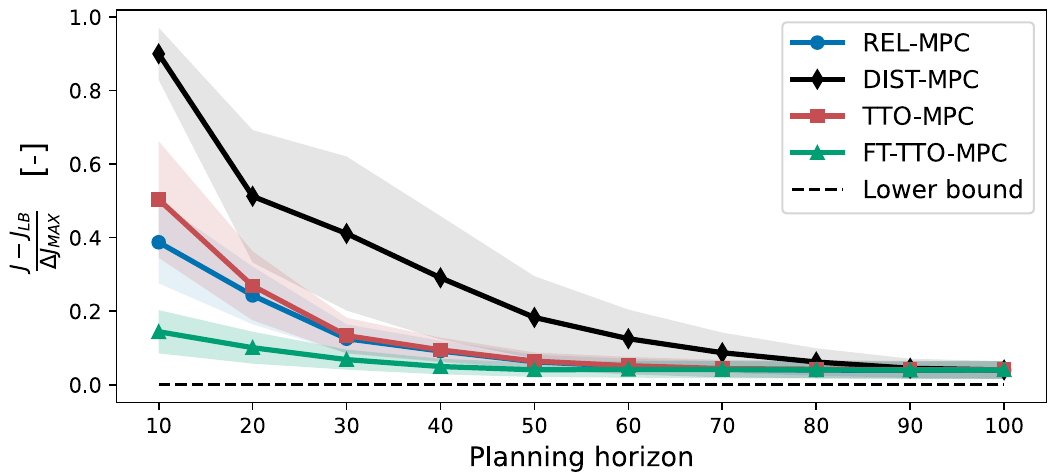}
    \caption{Average normalized cost increment with respect to the problem lower bound.}
    \label{fig:closed_loop_res}
\end{figure}
\begin{figure}[t]
\centering
    \includegraphics[width=0.85\linewidth]{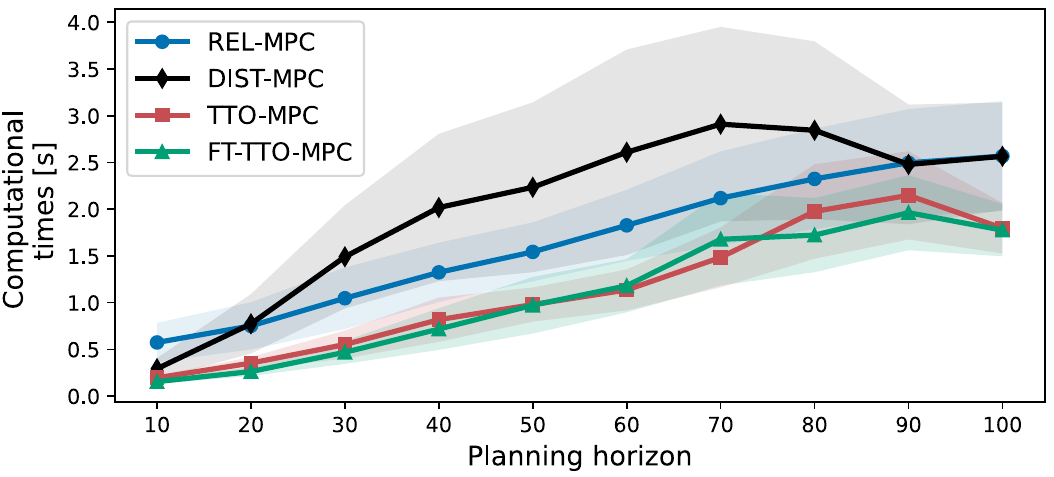}
    \caption{\black{Average maximum runtime per single MPC iteration. The runtime is defined as the total computation time needed to generate the warm-start and solve the OCP for the time horizon of interest.}}
    \label{fig:closed_loop_times}
\end{figure}
For each of the three simulation scenarios, we evaluate all methods on 500 problem instances uniformly distributed across the non-convexity factor.
Each trajectory is discretized in 100 timesteps and solved according to ten different planning horizons, ranging from 10 to 100 in increments of 10. 
To facilitate comparison across tasks, we normalize costs to the range between 0 and 1 by computing the normalized cost increment as $\frac{\texttt{cost - lowerbound}}{\texttt{maxcost - lowerbound}}$.
A normalized cost increment of 0 corresponds to a trajectory that matches the performance of the lower bound. 
A normalized cost increment of 1 corresponds to a trajectory that matches the worst performance observed across all planning horizons.
\black{Finally, the trajectory generation process of transformer models is initialized setting the parameters $\calR_1$ and $\calC_1$ as in Sec. \ref{sec:open_loop_res}.}
Aggregate results in Fig. \ref{fig:closed_loop_res} and \ref{fig:closed_loop_times} show how FT-TTO-MPC is able to substantially outperform all other approaches both in terms of cost and runtime\footnote{Computation times are from a Linux system equipped with a 4.20GHz processor and 128GB RAM, and a NVIDIA RTX 4090 24GB GPU.}.
In particular, Fig. \ref{fig:closed_loop_res} highlights a few key take-aways.
First, due to the short-sighted definition of the terminal cost, DIST-MPC exhibits the biggest performance degradation, showcasing the limits of heuristically defined terminal costs.
Second, TTO-MPC (trained solely open-loop data) is highly affected by the covariate shift problem, which severely impacts its closed-loop performance causing it to perform worse than REL-MPC.
Lastly, FT-TTO-MPC showcases the importance of fine-tuning transformer models for MPC, resulting in a significative advantage over all the other methods. 
In particular, the cost induced by FT-TTO-MPC remains stable across different planning horizons, achieving results with $H = 10$ that are on par with other methods with $H = 30$ and allowing for the solution of substantially smaller OCPs.

Results in \cref{fig:closed_loop_times} further highlight the runtime benefits of FT-TTO-MPC, which outperforms other non-learning-based methods across all planning horizons.
Crucially, by evaluating runtime results in the context of their cost performance from \cref{fig:closed_loop_res}, some relevant findings emerge.
In the context of REL-MPC, planning with $H=30$ is necessary to achieve performance comparable to the one obtained by FT-TTO-MPC with $H=10$.
Comparing the corresponding computational times, we see how the long-term guidance introduced by FT-TTO leads to approximately a $7$x reduction in runtime for the same performance.

\subsection{Real-world Experiments}
\begin{figure}[t]
\centering
    \includegraphics[width=0.85\linewidth]{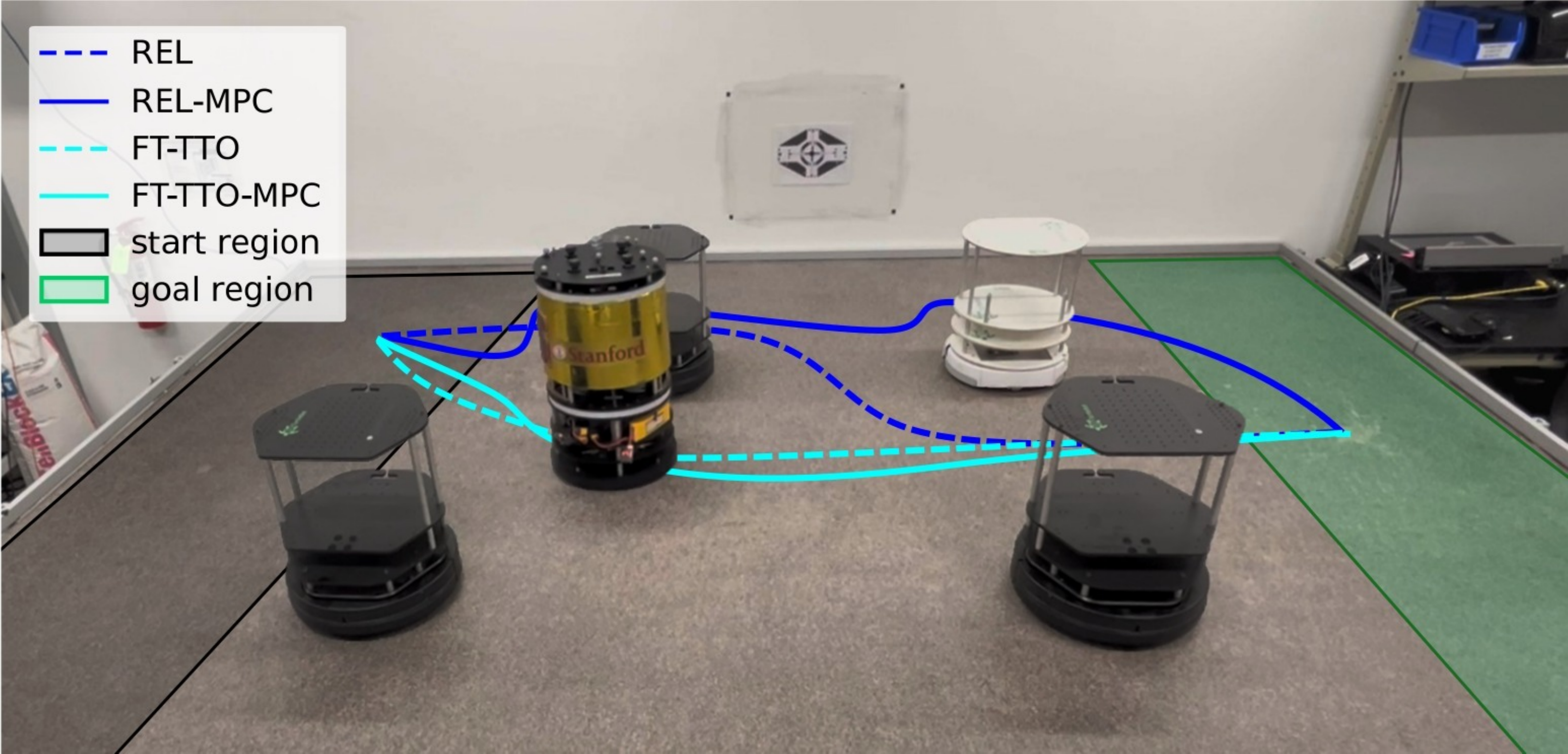}
    \caption{Qualitative representation of the experimental results obtained employing FT-TTO (cyan) and REL (blue) in the open-loop (dashed) and MPC (solid) settings. Videos: \href{https://transformermpc.github.io}{https://transformermpc.github.io}.}
    \label{fig:exp_tests}
\end{figure}
Finally, we perform experiments evaluating the real-world
effectiveness of our approach through the free flyer testbed described in Sec. \ref{sec:scenarios_description}.
Specifically, we use the cumulative firing time of the eight thrusters of the free flyer as a performance metric to compare FT-TTO and REL.
In the open-loop setting, we use a PID controller to track the solution obtained by warm-starting the SCP with FT-TTO and REL, respectively.
In the MPC framework, we directly control the free flyer using FT-TTO-MPC and REL-MPC with a planning horizon $H=10$.

\black{\cref{fig:exp_tests} and Tab. \ref{tab:exp_tests} show  qualitative and quantitative results for the free flyer experiments, focusing on highly non-convex OCPs}.
In the open-loop setting (O-L), FT-TTO facilitates the convergence to a solution with a reduced number of sharp turns (\cref{fig:exp_tests})---and thus, propellant consumption (Tab. \ref{tab:exp_tests})---clearly outperforming REL.
In the MPC framework, FT-TTO-MPC leads to a closed-loop execution that closely mimics the open-loop solution, limiting the increase in propellant consumption to approximately $20\%$, i.e., a value in line with the numerical analysis in simulation. 
On the other hand, REL-MPC results in a terminal cost that aggressively steers the free flyer toward the target, leading to significant deviations from the open-loop solution and abrupt turns in the proximity of the obstacles.
Furthermore, in our experiments, it was necessary to double the target acquisition time and halve the control frequency to have REL-MPC successfully reach the target while accounting for higher computation times and preventing collisions---a scenario we encountered consistently during experimentation with REL-MPC. 
As a result, the increased target acquisition times lead to an overall lower firing time, as shown in Tab. \ref{tab:exp_tests}.
Crucially, FT-TTO-MPC was the only MPC algorithm able to be deployed within real-world scenarios that did not require loosened time constraints.

{\renewcommand{\baselinestretch}{1} 
\setlength{\belowcaptionskip}{0mm}
\begin{table}[!t]
    \caption{Quantitative results of experimental tests shown in \cref{fig:exp_tests}.}
    \label{tab:exp_tests}
    \centering
    \resizebox{\columnwidth}{!}{
    \begin{tabular}{cc|ccc}
        \hline
        \hline
        \multicolumn{2}{c|}{\multirow{2}{*}{Method}} & Target acquisition & Frequency & Total firing \\
        \multicolumn{2}{c|}{} & time $[s]$ & $[Hz]$ & time $[s]$\\
        \hline
        \multirow{2}{*}{O-L} & REL & 40.0 & - & 99.70 \\
         & FT-TTO & 40.0 & - & \textbf{75.78}\\
        \hline
        \multirow{2}{*}{MPC} & REL-MPC & 80.0 & 1.25 & 83.76 \\
         & FT-TTO-MPC & \textbf{40.0} & \textbf{2.5} & 91.34\\
         \hline
         \hline
    \end{tabular}}
\end{table}}
\section{Conclusions}
\label{sec:conclusions}
Despite its potential, the application of learning-based methods for the purpose of trajectory optimization has so far been limited by the lack of safety guarantees and characterization of out-of-distribution behavior. 
At the same time, methods relying on numerical optimization typically lead to high computational complexity and strong dependency on initialization.
In this work, we address these shortcomings by proposing a framework to leverage the flexibility of high-capacity neural network models while providing the safety guarantees and computational efficiency needed for real-world operations.
We do so by defining a structure where a transformer model---trained to generate near-optimal trajectories---is used to provide an initial guess and a predicted target state as long-horizon guidance within MPC schemes.
This decomposition, coupled with a pre-training-plus-fine-tuning learning strategy, results in substantially improved performance, runtime, and convergence rates for both open-loop planning and model predictive control formulations.

\black{In future works, we plan to extend our formulation to handle multi-task stochastic optimization problems employing generalizable scene and input representations. Specifically, current limitations, such as unmodeled and out-of-distribution dynamics, will be addressed through tailored learning strategies (e.g. data augmentation and meta-learning of dynamics models) and structured treatment of uncertainties.}
Moreover, while we have investigated the impact of fine-tuning based on (iterative) supervised learning in this work, the exploration of other fine-tuning strategies (e.g., based on reinforcement learning) is a highly promising direction that warrants future work.
More generally, we believe this research opens several promising directions to leverage the power of learning-based methods for trajectory optimization within safety-critical robotic applications.

                                  

\bibliographystyle{IEEEtran}
\small
\bibliography{ASL_Bib, main}
\end{document}